\documentclass[sigconf,nonacm]{acmart}
\usepackage{makecell}
\newtheorem{rem}{\bf{Remark}}
\usepackage{multirow}
\AtBeginDocument{%
  }

\acmConference[ICAIF'25]{6th ACM International Conference on AI in Finance}{November 15--17,2025}{Singapore}

\begin{document}

\title{FinDPO: Financial Sentiment Analysis for Algorithmic Trading through Preference Optimization of LLMs}

\author{Giorgos Iacovides}
\email{giorgos.iacovides20@imperial.ac.uk}
\affiliation{%
  \institution{Imperial College London}
  \city{London}
  \country{UK}
}
\author{Wuyang Zhou}
\email{wuyang.zhou19@imperial.ac.uk}
\affiliation{%
  \institution{Imperial College London}
  \city{London}
  \country{UK}
}

\author{Danilo Mandic}
\email{d.mandic@imperial.ac.uk}
\affiliation{%
  \institution{Imperial College London}
  \city{London}
  \country{UK}
}

\renewcommand{\shortauthors}{Iacovides et al.}

\begin{abstract}
Opinions expressed in online finance-related textual data are having an increasingly profound impact on trading decisions and market movements. This trend highlights the vital role of sentiment analysis as a tool for quantifying the nature and strength of such opinions. With the rapid development of Generative AI (GenAI), supervised fine-tuned (SFT) large language models (LLMs) have become the \textit{de facto} standard for financial sentiment analysis. However, the SFT paradigm can lead to memorization of the training data and often fails to generalize to unseen samples. This is a critical limitation in financial domains, where models must adapt to previously unobserved events and the nuanced, domain-specific language of finance. To this end, we introduce \textit{FinDPO}, the first finance-specific LLM framework based on post-training human preference alignment via Direct Preference Optimization (DPO). The proposed FinDPO achieves state-of-the-art performance on standard sentiment classification benchmarks, outperforming existing supervised fine-tuned models by 11\% on the average. Uniquely, the FinDPO framework enables the integration of a fine-tuned causal LLM into realistic portfolio strategies through a novel ‘logit-to-score’ conversion, which transforms discrete sentiment predictions into continuous, rankable sentiment scores (probabilities). In this way, simulations demonstrate that FinDPO is the first sentiment-based approach to maintain substantial positive returns of 67\% annually and strong risk-adjusted performance, as indicated by a Sharpe ratio of 2.0, even under realistic transaction costs of 5 basis points (bps). 
 
\end{abstract}




\maketitle

\section{Introduction}
Financial sentiment analysis refers to the quantification of opinions present in unlabeled textual data, and aims to categorize whether the overall perspective is positive, negative, or neutral - the so called valence. Given the rapidly increasing volume of financial sentiment-related textual information, and its readily available nature, it comes as no surprise that it has a significant influence on financial markets. This, coupled with the key role of algorithmic trading in quantitative finance, has highlighted the need for reliable and actionable AI models which are trained on such vast, multimodal data streams. Of particular interest to sentiment analysis is generative AI (GenAI), owing to its ability to recognize sentiment from textual sources such as news articles, earnings calls, financial reports, and other non-numerical data. The capacity of GenAI to operate on large-scale information sources makes it possible to capture macroscopic trends, promising a competitive edge in stock price prediction \cite{stock_prices} and the development of effective and robust trading strategies \cite{trad_strats}. \par 
Despite unquestionable conceptual benefits, the diverse, nuanced, and domain-specific nature of financial text poses significant challenges when it comes to accurate and actionable sentiment extraction. This underscores the need for context-aware sentiment analysis and highlights the non-trivial obstacles of employing natural language processing (NLP) in financial applications. Large language models (LLMs) have emerged as powerful tools for addressing these challenges, particularly when fine-tuned on labeled financial datasets through post-training techniques. Indeed, supervised fine-tuning (SFT) methods, particularly instruction tuning, which utilize the remarkable capability of LLMs to comprehend and generate human-like text, have become the \textit{de facto} standard for enhancing the performance of pre-trained LLMs on financial sentiment classification tasks. Despite the initial success, recent work has shown that the SFT paradigm can lead to memorization of the training data while struggling to generalize to unseen samples \cite{rl_sft}.  This limitation is particularly critical in financial domains, where the ability to generalize to previously unseen events is essential for robust algorithmic trading strategies. It also stands in stark contrast to the behavior of human financial analysts, who are able to extrapolate from limited prior information and apply domain knowledge to assess sentiment in new and unexpected market conditions. \par
\begin{figure*}[!t]
\centering
    \includegraphics[width=0.95\textwidth]{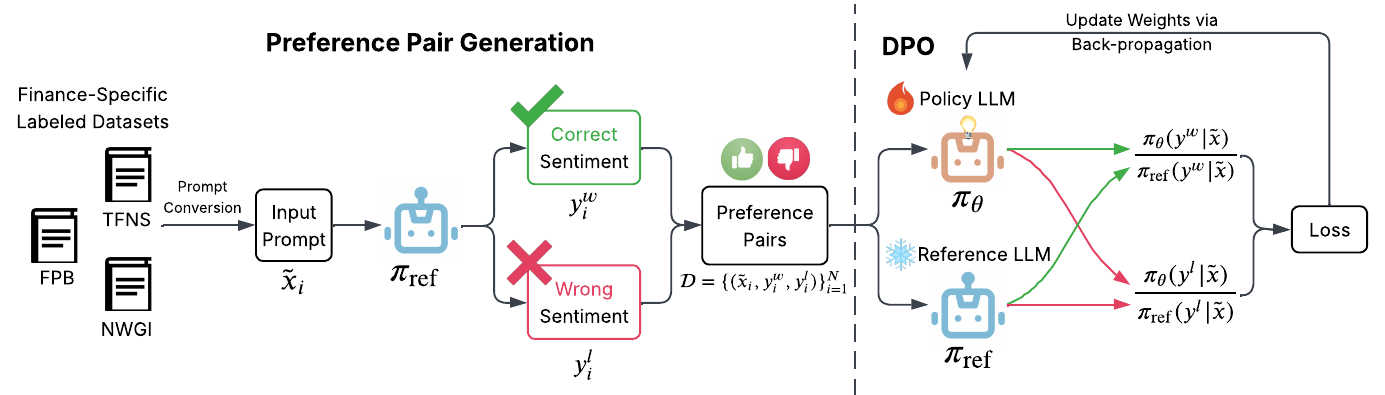}
    \caption{FinDPO training principle. Preference pairs are first generated from finance-specific labeled datasets. These preference pairs are then fed into the policy LLM whose weights are updated and the reference LLM whose weights are frozen. The weights of the policy LLM are updated by backpropagating the DPO loss, computed as the difference between the log-ratios of the policy and reference model probabilities for the preferred (correct) and dispreferred (incorrect) sentiment labels, as detailed in Equation~\ref{dpo_loss}. Intuitively, these updates encourage the policy model to increase the likelihood of selecting preferred sentiment labels while decreasing the likelihood of selecting dispreferred ones.}
    \label{fig:Flowchart}
\end{figure*}
To address these issues, we set out to answer:
\begin{itemize}
    \item Can we develop a financial sentiment analysis framework which goes beyond supervised fine-tuning and aligns large language models with human preferences as a means to enhance algorithmic trading?
    \item Can this be achieved in a way which does not require extensive computational resources, typically required by LLM post-training techniques, thus making the approach operational on standard computational resources?
\end{itemize}
To this end, we introduce the first post-training human preference alignment technique for financial sentiment analysis, based on Direct Preference Optimization (DPO) \cite{dpo}. This allows us to leverage human-alignment methods, which have demonstrated improved generalization to out-of-distribution inputs in other NLP tasks \cite{rl_sft}, such as translation and summarization, in the financial domain. In particular, our solution, termed \textit{FinDPO}, combines human preference optimization with a pre-trained LLM (specifically, Llama-3-8B Instruct \cite{llama3}) for financial sentiment analysis, enabling effective training on a small corpus of specialized, labeled, and publicly available financial news datasets. In this way, FinDPO achieves its ultimate goal of enhancing financial sentiment analysis through parameter-efficient DPO, while also minimizing computational resource requirements.

The main contributions of this work are therefore:
\begin{itemize}
    \item We propose \textit{FinDPO}, the first LLM aligned with human preferences specifically for financial sentiment analysis. 
    \item The FinDPO approach does not require multiple high-end GPUs and can operate on standard computational resources. By leveraging the pre-trained Llama-3-8B model and employing parameter-efficient techniques within the DPO training, the computational demands typically associated with preference alignment approaches are dramatically reduced.
    \item The proposed framework enables, for the first time, causal LLMs to be integrated into long-short portfolio strategies by converting their discrete sentiment labels into continuous scores. This allows for evaluation through finance-specific, real-world performance metrics.
    \item Simulations demonstrate that FinDPO achieves substantial improvements in performance over existing state-of-the-art SFT models in both standard classification benchmarks and real-world financial metrics.
    
\end{itemize}
\begin{rem}
Finance-specific causal LLMs have shown the most
promising performance in sentiment classification tasks, however, they have not yet adequately addressed the domain of portfolio construction. By moving beyond traditional supervised fine-tuning toward human-aligned LLMs, and enabling the benchmarking of causal LLMs through algorithmic trading metrics, the contributions of our work promise to establish a paradigm shift in financial sentiment analysis. 
\end{rem}
\section{Related Work}
\textbf{Post-Training Techniques in Large Language Models.} Even though LLMs trained on massive textual corpora possess remarkable general capabilities, their performance on downstream tasks can be significantly improved through post-training fine-tuning techniques. Post-training approaches typically fall into two categories: i) supervised fine-tuning (SFT), including instruction tuning, which adapts a pre-trained LLM using labeled input-output pairs, and ii) preference optimization. Despite success, instruction tuning remains limited in its ability to capture nuanced user preferences. Consequently, human preference optimization was introduced, initially through Reinforcement Learning from Human Feedback (RLHF) \cite{rlhf}. This approach typically involves first training a reward model on a dataset of human preferences, and then fine-tuning the language model to maximize this reward using reinforcement learning algorithms, most commonly REINFORCE \cite{reinforce}, Proximal Policy Optimization (PPO) \cite{ppo}, or their variants \cite{variants}. While RLHF has demonstrated superior performance compared to instruction tuning, it is often computationally expensive and can be unstable due to the challenges of reward modeling and the complexity of the RL optimization process. \par
To mitigate these challenges, Direct Preference Optimization (DPO) \cite{dpo} has been proposed as a simpler, RL–free, and more stable alternative for training language models from preference data. Importantly, while DPO implicitly optimizes the same objective as RLHF (maximizing reward subject to a KL-divergence constraint), it achieves this in a direct and more tractable manner, maintaining ease of implementation and training stability. \newline  \newline  
\textbf{Sentiment Analysis with Large Language Models.} The advent of transformer-based models for financial sentiment analysis began with FinBERT \cite{araci2019finbert}, a variant of BERT fine-tuned on financial text, which demonstrated promising performance for sentiment classification tasks in the financial domain. However, FinBERT suffers from limitations such as insensitivity to numerical values, while due to its relatively small size (110 million parameters) its classification accuracy deteriorates with increased sentence complexity \cite{finbert_complexity}. More recently, Instruct-FinGPT \cite{zhang2023instructfingpt} and FinGPT \cite{fingpt} models have adopted instruction tuning to enhance model performance, by leveraging the Llama-7B and Llama-2-13B models, respectively, as their base model. While these models represent a shift towards more powerful and generalizable LLMs, FinGPT is not specifically optimized for financial sentiment analysis. Furthermore, both models are limited to predicting sentiment valence (i.e., positive, negative, or neutral) but lack the ability to quantify the strength of a sentiment class, an essential parameter for portfolio construction. \par 
In contrast, FinLlama \cite{finllama} combines SFT with the Llama-2-7B model and introduces a classification head at the output of the LLM to produce continuous sentiment scores. 
While this modification enables the sentiment signal to be integrated directly into portfolio construction workflows, it also shifts the primary objective of the model from next-token prediction to classification. This restricts the applicability of more advanced post-training techniques that depend on the generative capabilities of language models. \par
To address these limitations we propose FinDPO, the first finance-specific LLM for sentiment analysis based on post-training human preference alignment, rather than SFT. Our model is motivated by the fact that preference optimization has shown superior performance and generalization over SFT, particularly in tasks that involve capturing subtle human preferences expressed in nuanced language \cite{dpo, rlhf}, a characteristic that also underpins our task of financial sentiment classification.

\section{Preliminaries}
Given dataset $\mathcal{D}$, the objective of DPO is to minimize

\begin{equation} \label{dpo_loss}
\small
\mathcal{L}_{\mathrm{DPO}} = -\mathbb{E}_{(\tilde{x}, y^w, y^l)} \left[ \log \sigma \left( \beta \cdot \left( \log \frac{\pi_{\theta}(y^w|\tilde{x})}{\pi_{\mathrm{ref}}(y^w|\tilde{x})} - \log \frac{\pi_{\theta}(y^l|\tilde{x})}{\pi_{\mathrm{ref}}(y^l|\tilde{x})} \right) \right) \right],
\end{equation}
where $\sigma$ denotes the nonlinear sigmoid function, and $\beta$ controls the deviation from the base reference policy, $\pi_{\mathrm{ref}}$, namely the initial reference LLM. \par
The typical workflow of DPO is outlined below:
\begin{enumerate}
    \item For each prompt, $\tilde{x}$, sample completions $y_1, y_2 \sim \pi_{\mathrm{ref}}(\cdot | \tilde{x})$. 
    \item Collect human preference labels indicating which of $y_1$ or $y_2$ is preferred, yielding an offline dataset of preference pairs $\mathcal{D} = \{(\tilde{x}_i, y_i^w, y_i^l)\}_{i=1}^N$.
    \item Optimize the policy (trainable) model, $\pi_\theta$, to minimize the DPO loss given the reference (frozen) model, $\pi_{\mathrm{ref}}$, the dataset, $\mathcal{D}$, and the desired hyperparameter, $\beta$.
\end{enumerate} 
Intuitively, DPO updates the parameters of the policy model, $\pi_\theta$, to increase the likelihood of preferred responses, $y^w$, and decrease the likelihood of dispreferred responses, $y^l$. The magnitude of these updates is modulated by how strongly $\pi_\theta$ disagrees with human preferences, scaled by $\beta$. This dynamic weighting stabilizes training and prevents model collapse.

\section{Methodology}
Our work aims to leverage the expressive power and contextual understanding of general-purpose LLMs and adapt them for sentiment analysis applications. This is achieved by applying human preference alignment, using DPO, to Llama-3-8B-Instruct, which designates our $\pi_{\mathrm{ref}}$ model, and aligning it on a finance-specific corpus. The effectiveness of our approach is evaluated in two ways. First, we assess the performance of the model relative to other finance-specific LLMs using standard classification-focused benchmarks, such as the weighted F1 score. For rigour, our model is further evaluated through a set of benchmarks that closely align with real-world portfolio construction — the ultimate goal of sentiment analysis.
\subsection{Training Pipeline of FinDPO}
Pre-trained LLMs offer a range of capabilities such as reasoning, translation, summarization, and text generation, however, they often struggle when applied directly to domain-specific tasks such as financial sentiment analysis. This limitation is particularly pronounced in the finance domain, where nuanced language, speculative narratives, and the extensive length of financial news articles pose significant challenges for general-purpose models. \par 
To address these challenges, our work revisits the first principles of LLMs and leverages human preference alignment to adapt them for financial sentiment analysis. More specifically, we utilize DPO, which operates directly on pairwise preferences constructed from ground-truth sentiment labels and model predictions, enabling the model to learn fine-grained distinctions in financial sentiment while preserving its generative capabilities. Through this process, the proposed FinDPO model produces sentiment predictions for three classes: positive, negative, or neutral.
\subsubsection{Training Datasets.} The training data consisted of three publicly available labeled financial news datasets, obtained from HuggingFace: the Financial PhraseBank (FPB) dataset \cite{malo2014good}, the Twitter Financial News dataset \cite{GPT_news}, and the GPT-labeled Financial News dataset \cite{twitter_sent}. This resulted in a total of 32,970 labeled samples, of which 80\% were used for training and the remaining 20\% for testing. The datasets are summarized below.
\begin{itemize}
    \item \textbf{Financial PhraseBank (FPB) Dataset.} The FPB dataset consists of 4,840 samples which were randomly extracted from financial news articles. In order to ensure high quality annotation, the samples were annotated by 16 experts with backgrounds in finance and business. Each sample was annotated with one of the three labels: positive, negative, and neutral.  
    \item \textbf{Twitter Financial News Sentiment (TFNS).} The TFNS dataset includes 11,930 tweets with content from the financial domain. Each tweet was annotated as positive, negative, and neutral. 
    \item \textbf{GPT-labelled Financial News (NWGI).} The NWGI dataset consists of 16,200 financial news articles. Each article was annotated with one of the five labels: strongly negative, mildly negative, neutral, mildly positive, and strongly positive. To align this dataset with the three-class structure of the other datasets, the strongly and mildly negative classes were combined into a single negative class, and similarly, the strongly and mildly positive classes were combined into a single positive class.  
\end{itemize}

As DPO requires \textit{preference pairs} rather than \textit{class labels}, we converted those three datasets into synthetic pairwise preference data. Each raw text input, $x_i$, was first converted into its instruction format, $\tilde{x}_i$, compatible with the instruction-tuned model. For each sample, we set the preferred response, $y^w_i$, to the ground-truth sentiment label. To obtain the dispreferred response, $y^l_i$, we prompted the reference model, $\pi_{\mathrm{ref}}$, with $\tilde{x}_i$ to generate sentiment predictions. If the prediction matched the ground-truth sentiment label, we randomly sampled a different incorrect label as $y^l_i$ to avoid bias in selecting the dispreferred response. If the prediction was incorrect, we used the predicted label as $y^l_i$ in order to guide the model away from its own mistakes and reinforce correct sentiment predictions. This resulted in a dataset of preference pairs $\mathcal{D} = \{(\tilde{x}_i, y^w_i, y^l_i)\}_{i=1}^N$, which was used for DPO training.
\subsubsection{Model Training.} The proposed FinDPO model was first initialized with the Llama-3-8B-Instruct model, which serves as the base reference policy, $\pi_{\mathrm{ref}}$, followed by DPO alignment over 5 epochs. The training process employed the AdamW optimizer \cite{Loshchilov2017FixingWD}, as it effectively decouples the weight decay from the optimization steps and leads to more stable convergence. The initial learning rate was deliberately set to a small value as the Llama-3 model is already pre-trained on a large corpus of data, whilst the warm-up ratio and weight decay served as key regularization techniques to prevent overfitting, a crucial aspect given the limited size of our DPO training dataset. \par 
Furthermore,  Low-Rank Adaptation (LoRA) \cite{lora} was integrated into the DPO training with a rank $r=16$, scaling factor $\alpha=16$, and dropout rate of 0.05, in order to minimize the number of trainable parameters while maintaining high end performance. Through the LoRA implementation, the number of trainable parameters was set to 41.9M, amounting to just 0.52\% of the total number of parameters in the base model. This made it possible for our training process to be \textbf{implemented on a single A100 (40 GB) GPU}, thus avoiding the need for excessive computational resources. The full training procedure was completed in 4.5 hours on an A100 (40 GB) GPU . \par
The entire training pipeline, including the conversion of raw training data into a dataset of preference pairs, $\mathcal{D}$, and the subsequent DPO alignment of the base reference policy, $\pi_{\mathrm{ref}}$, on $\mathcal{D}$ to produce our FinDPO model, is given in Figure \ref{fig:Flowchart}.
\subsection{Proposed Framework for Sentiment-Driven Portfolio Construction} \label{portfolio_construction} 
\begin{figure}[htbp] 
\centering 
\includegraphics[width=0.48\textwidth]{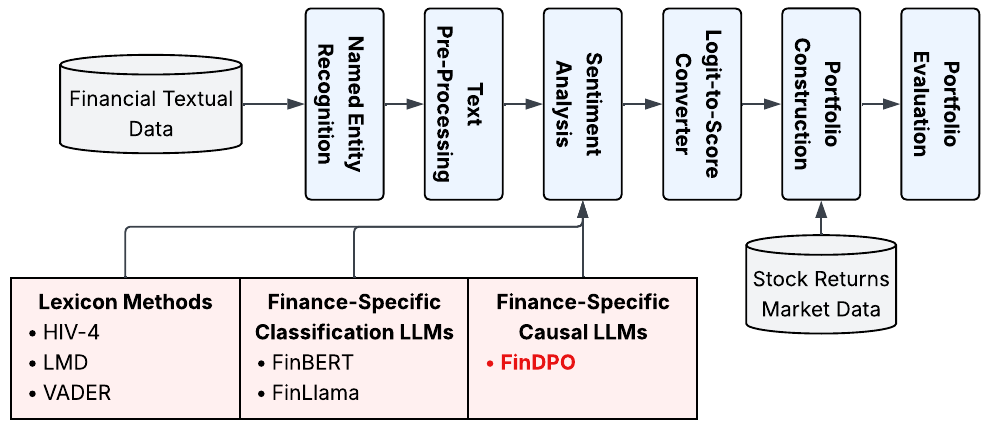}
\caption{Proposed framework for our sentiment-driven portfolio construction. The `logit-to-score' converter is only required for the finance-specific causal LLMs.}
\Description{Framework for sentiment analysis. Each stage is fully described in the text}
\label{fig:project_framework}
\end{figure}

Once our FinDPO model was established, we followed the framework illustrated in Figure~\ref{fig:project_framework}. The objective of our framework was to evaluate the performance of FinDPO against other established sentiment analysis methods using finance-specific real-world metrics, offering a more practical assessment of model utility for portfolio construction.\par 
 \par 
\noindent \textbf{Data Collection.} Both textual and market data were collected to support the construction of appropriate long-short (L/S) portfolios. For the textual data, 204,017 financial news articles dating between February 2015 and June 2021 were gathered from reputable online sources, including Reuters, The Motley Fool, and MarketWatch. These sources were chosen for their reliability, reputation, low bias, and focus on major corporations. For the same period, the corresponding financial market data were obtained from Yahoo Finance, comprising daily stock returns for the 500 companies in our Investable Universe (S\&P 500). This resulted in 1,672 days of stock return data per company. \par 
\noindent \textbf{Named Entity Recognition.} To ensure that news articles are accurately linked to the correct organizational entity, each article must be associated with at least one relevant stock \cite{one_stock}. This step reduces the likelihood of irrelevant articles being connected to a particular stock \cite{irrelevant_articles}. For robustness, in our study we employed the BERT-base-NER model \cite{bert_base_ner}, which is capable of recognizing four types of entities: location, organization, person, and miscellaneous, and provides a confidence score for each identified entity. For each article in our corpus, if the confidence score for the entity associated with the company exceeded 98\%, the article was retained; otherwise, it was discarded. Table \ref{tab:no_of_articles} presents the results of applying the NER filtering on the initially scraped dataset in terms of the number of articles retained.  Observe that this process reduced the total number of articles by 24.1\%. The effect was similar for MarketWatch and The Motley Fool, with both exhibiting reductions of around 25\%, while Reuters experienced a smaller decrease of approximately 6.3\%.
\begin{table}[h!]
\centering
\caption{Total number of articles obtained per source before and after NER filtering, with percentage reduction shown in brackets.}
\resizebox{\linewidth}{!}{\begin{tabular}{lcc}
\hline
\textbf{News Source} & \textbf{No. of articles pre processing} & \textbf{No. of articles post NER filtering} \\ \hline
MarketWatch & 309,187 & 236,214 (23.6\%)\\
Reuters & 38,141 & 35,741 (6.3\%)\\
The Motley Fool & 205,270 & 147,413 (28.2\%)  \\
\hline
\textbf{Total} & \textbf{552,598} & \textbf{419,368} (\textbf{24.1\%}) \\
\hline
\end{tabular}}
\label{tab:no_of_articles}
\end{table} 
\par 
\noindent \textbf{Text Pre-Processing}: Each news article was represented using a bag-of-words approach, and the following steps were subsequently performed: a) Tokenization, b) Stop-Word Removal, c) Lemmatization, d) (Lower) Case Normalization, and e) Feature Selection. For the Feature Selection step, the frequency of each word was used as the primary feature, particularly in the lexicon-based approaches. \par
\noindent \textbf{Sentiment Analysis.} In total, five sentiment analysis methods were used as baselines for comparison with our FinDPO model. These consisted of lexicon-based approaches and SFT classification LLMs, including the existing SOTA FinLlama \cite{finllama}. Among the lexicon-based methods, LMD \cite{lmd} and HIV-4 \cite{hiv4} were implemented using the pysentiment2 Python library, while VADER \cite{vader} was implemented using the NLTK library. For SFT classification models, we evaluated FinBERT \cite{araci2019finbert} and FinLlama \cite{finllama}, obtained through HuggingFace and utilized via the Transformers library. \par
Despite exhibiting the strongest performance on standard classification focused benchmarks (as shown in Table~\ref{tab:results_ml}), fine-tuned \textbf{causal} LLMs are inherently limited to generating discrete sentiment labels (i.e., positive, negative, neutral) and do not capture the sentiment strength. However, understanding how strongly a sentiment is expressed is crucial for portfolio construction, where assets must be ranked based on sentiment magnitude rather than simply classified into categories. To overcome this limitation and demonstrate the feasibility of using causal LLMs for portfolio construction, we propose a novel `\textit{logit-to-score}' converter within our FinDPO framework, which enables sentiment strength ranking by leveraging the internal representations of the model. 
\begin{rem}
To make causal LLMs amenable to portfolio construction, we extract the logits of the model corresponding to the first generated token, which represents the predicted sentiment class, and apply a softmax function over the logits associated with the predefined sentiment classes (i.e., `positive', `negative', or `neutral'). This results in a normalized probability distribution over the sentiment labels, which we interpret as sentiment scores suitable for ranking assets in a portfolio.
\end{rem}
Related to our FinDPO model, prior work has shown that preference aligned LLMs can exhibit overconfidence \cite{llm_confidence}. We have also observed this effect in our model, whereby it consistently assigned a probability of 1.0 to the predicted valence class and 0.0 to the other two valence classes. To mitigate this and produce more meaningful probabilistic sentiment scores, we employed temperature scaling \cite{temp_scaling}, a post-hoc calibration method. The temperature parameter $T$ was optimized on the training set by minimizing the negative log-likelihood (NLL). This calibration was performed independently of the financial article corpus used for portfolio construction, ensuring no data leakage. \par 
The considered methods were evaluated on every article within each corpus for a given company. In cases where multiple articles were published on the same day for a given company, the average sentiment for that day was calculated as
\begin{equation}
S_t=\frac{1}{N_t}\sum_{i=1}^{N_t} S_{it}
\end{equation}
Here, $S_t$ represents the average sentiment for the t-th day, $N_t$ denotes the number of news articles published on that same t-th day for a given company, while $S_{it}$ designates the sentiment strength of the i-th news article on a particular t-th day. The daily sentiment outputs for each company were merged to arrive at the final sentiment data that were utilized as a parameter in the portfolio construction stage. \par 
\noindent \textbf{Portfolio Construction.} Once the sentiment for each method was defined for every company, the long-short portfolio was constructed. We used the sentiment as a parameter to determine which companies should be in a long or a short position, aiming to maximize returns from both positions. The long-short portfolio was constructed using the following procedure:
\begin{itemize}
    \item \textit{Define the Investable Universe:} Even though the S\&P 500 comprises 500 companies, the financial textual data collected did not contain articles associated to some of the companies for the test period of February 2015 to June 2021. Consequently, 417 companies were considered. 
    \item \textit{Define the long and short position}: The sentiment signal obtained from each of the six methods was used to construct six distinct portfolios. For each method, companies were ranked daily according to their sentiment. Companies that did not have sentiment data on a particular day were omitted from the ranking. As the daily sentiment score for each company ranges between -1 and 1, those with the highest positive sentiment were placed in a long position, while those with the strongest negative sentiment were placed in a short position.
    \item \textit{Allocation:} An equally-weighted portfolio strategy was considered in our portfolio construction as this strategy is  mostly utilized by hedge funds \cite{hedge_funds_strat}. Similar to \cite{finllama}, the percentage of companies in a long and short position was fixed at 35\%. Consequently, the top 35\% of companies in terms of performance were allocated to long positions, while the bottom 35\% were allocated to short positions.  
    \item \textit{Determine daily returns:} The daily return for each company that was held in a long or short position was obtained by the market data on that particular day. The average daily return of companies that were held in a long position, $r_{Long}$, was defined as  
    \begin{equation} 
    r_{Long}=\frac{1}{N_{Long}}\sum_{i=1}^{N_{Long}} r_{Long}(i)
    \end{equation}
    Similarly, the average daily return of companies that were held in a short position, $r_{Short}$, was defined as
    \begin{equation}
    r_{Short}=\frac{1}{N_{Short}}\sum_{i=1}^{N_{Short}} r_{Short}(i)
    \end{equation}
    For each particular day, the number of companies that were held in either a long position ($N_{Long}$) or a short position ($N_{Short}$) were equal. Consequently, the total portfolio return on a particular day is the difference between the daily long return, $r_{Long}(i)$, and daily short return, $r_{Short}(i)$, and is given by
    \begin{equation}
    r_{daily}(i)=r_{Long}(i) - r_{Short}(i)
    \end{equation}
\end{itemize} \par 
\noindent \textbf{Portfolio Evaluation.} The performance of the portfolio constructed using the proposed model was assessed against the portfolios constructed using the other sentiment methods. We considered: 
\begin{itemize}
    \item \textit{Profitability}, via cumulative returns, $r_{\text{cum}}$, and annualized returns, $R_p$, defined as 
    \begin{equation}
    r_{\text{cum}} = \sum_{i=1}^{N} r_{\text{daily}}(i)
    \end{equation}
    \begin{equation}
    R_p = \frac{1}{N} \sum_{i=1}^{N} r_{\text{log}}(i) \cdot  252
    \end{equation}
    \item \textit{Risk-adjusted performance}, through annualized Sharpe,  $S_a$, Sortino, $S_o$, and Calmar, $C_r$, ratios, defined as 
    \begin{equation}
    S_a = \frac{R_p - R_f}{\sigma_p}
    \end{equation}
    \begin{equation}
    S_o = \frac{\bar{r}_{\text{simple}} - R_f}{\sigma_d} \cdot \sqrt{252}
    \end{equation}
    \begin{equation}
    C_r = \frac{(1 + \bar{r}_{\text{simple}})^{252} - 1}{\text{MDD}}
    \end{equation}
\end{itemize}
Here, \( N \) is the total number of trading days, totaling 1,672, \( r_{\text{log}}(i) \) represents the daily logarithmic return, and \( \bar{r}_{\text{simple}} \) denotes the average daily simple return. The symbol \( R_f \) is the annualized risk-free rate of return, \( \sigma_p \) is the annualized volatility, \( \sigma_d \) is the downside deviation of daily returns, and MDD is the maximum drawdown. The constant 252 corresponds to the number of business days in a calendar year. The risk-free return, $R_f$, typically represents the yield of the 10-Year Treasury Note; however, due to its prolonged low yield during the analyzed period \cite{low_yield}, a 0\% rate is commonly used and was adopted in our analysis. 
\section{Experimental Results}
We evaluated the effectiveness of our proposed FinDPO framework through both machine learning and financial performance assessments. The  machine-learning evaluation focused on standard sentiment classification benchmarks to measure predictive performance. The financial evaluation examined the real-world utility of FinDPO by integrating its outputs into an algorithmic trading strategy and analyzing portfolio performance under both idealized and realistic market conditions. 
\subsection{Evaluation via Classification Metrics}
The performance of our FinDPO model was assessed by first evaluating it on the test splits of the three datasets used for training, which are considered the standard benchmarks in financial sentiment analysis, using the weighted F1 score. Our model was compared against classical lexicon-based methods, as well as FinBERT and Instruct-FinGPT. In addition, we compared FinDPO against the state-of-the-art models in financial sentiment analysis, namely FinLlama and FinGPT v3.3. To further evaluate the effectiveness of DPO relative to instruction tuning -- the most prominent approach for improving financial sentiment classification in general-purpose LLMs (as demonstrated in FinGPT v3.3) -- we compared FinDPO to its instruction-tuned counterpart, using the same base model, which we refer to as \textit{FinSFT}.
\begin{table}[h!]
\centering
\caption{Weighted F1 scores of the evaluated sentiment-based methods on the FPB, TFNS, and NWGI financial sentiment datasets. HIV-4, VADER, and LMD are lexicon-based baselines. FinBERT and FinLlama are SFT classification LLMs, while Instruct-FinGPT, FinGPT v3.3 and FinSFT are instruction-tuned causal LLMs. FinDPO is our proposed DPO-aligned causal LLM. FinBERT is excluded from evaluation on FPB due to data leakage, as it was originally trained on the full FPB dataset.}
\tiny
\resizebox{\linewidth}{!}{\begin{tabular}{lcccc}
\hline
\textbf{Model} & \textbf{FPB} & \textbf{TFNS} & \textbf{NWGI} & \textbf{Average} \\ \hline
HIV-4 & 0.357 & 0.401&  0.384 & 0.385  \\
VADER & 0.536 & 0.518 & 0.462 &  0.491 \\
LMD & 0.546  & 0.572 & 0.440  & 0.498 \\
FinBERT & --- & 0.733 & 0.538 & 0.611\\
FinLlama & 0.707 & 0.904 & 0.538 &  0.679\\
Instruct-FinGPT & 0.777 & 0.828 & 0.583 & 0.690\\
FinGPT v3.3 & 0.879 & 0.903 & 0.643 & 0.762 \\
\textit{FinSFT} & 0.829 & 0.850 & 0.708 & 0.771 \\
\textbf{FinDPO (Ours)} & 0.865 & 0.872 & 0.833 & \textbf{0.846} \\
\hline
\end{tabular}}
\label{tab:results_ml}
\end{table} \par 
Table \ref{tab:results_ml} shows that FinDPO achieves the highest average performance across all three benchmark datasets, with a weighted F1 score of 0.846, outperforming FinGPT v3.3, the state-of-the-art model in financial sentiment classification, by 11\%. Notably, FinSFT delivers performance comparable to FinGPT v3.3, despite being based on the newer Llama-3-8B model rather than Llama-2-13B, suggesting that instruction-tuned models have reached a performance ceiling. This comparison highlights that the superior performance of FinDPO is not due to a more powerful base model, but rather to the use of our sentiment-specific DPO framework, which aligns model outputs with ground truth labels more effectively than instruction tuning in the context of financial sentiment classification.

\begin{rem}
The introduced DPO not only encourages the LLM to increase the likelihood of preferred sentiment labels but also explicitly penalizes the selection of dispreferred ones, resulting in better generalization to unseen samples compared to SFT methods which solely optimize for correct predictions. 
\end{rem}

\begin{figure*}[!t]
\centering
    \includegraphics[width=0.95\textwidth]{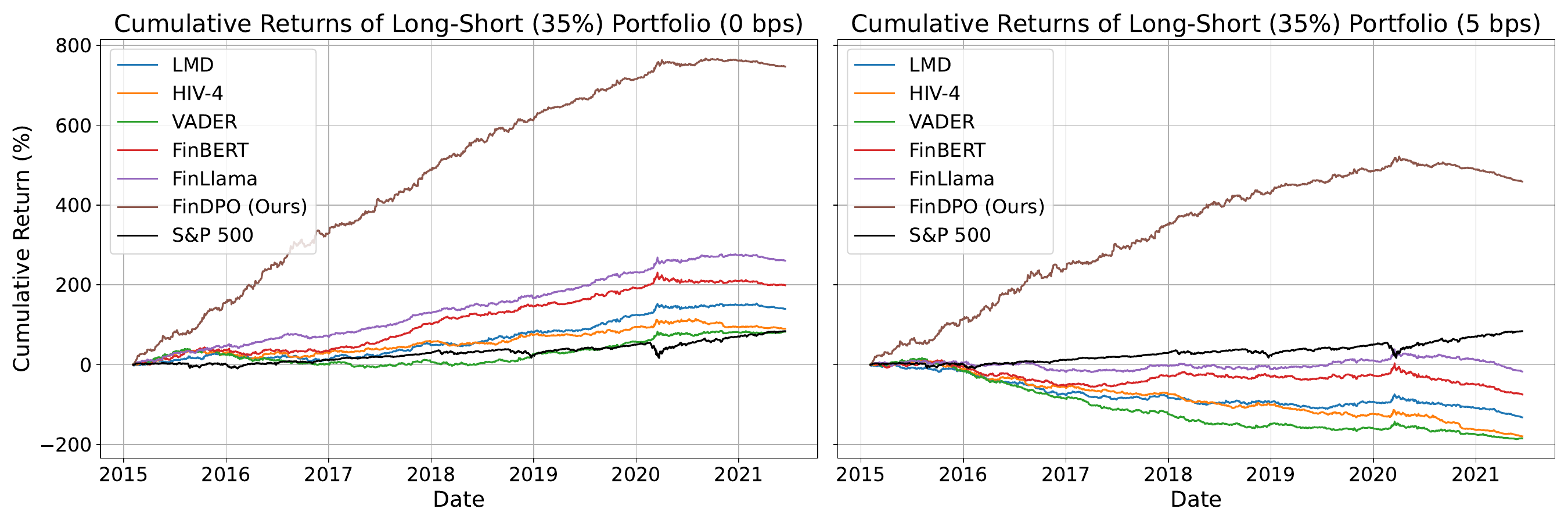}
    \caption{Cumulative returns of sentiment-based long-short portfolios at 0 and 5 bps transaction costs, compared against the S\&P 500 benchmark.}
    \label{fig:results}
\end{figure*}
\subsection{Evaluation via Real-World Financial Metrics}
We next assessed the performances of the six portfolios which were constructed as described in Section \ref{portfolio_construction}, without accounting for transaction costs. These are illustrated in the left panel of Figure \ref{fig:results} and Table \ref{tab:performance_0bps}. Observe that our proposed method, FinDPO, significantly outperforms the existing state-of-the-art model in sentiment-based portfolio construction, FinLlama, achieving a 187\% improvement in cumulative returns. Critically, FinDPO also demonstrates superior risk-adjusted performance, with an increase of 1.1 in the Sharpe ratio, indicating a more favorable return-to-volatility trade-off. Furthermore, FinDPO achieves the highest Sortino and Calmar ratios (6.05 and 11.94, respectively), reflecting improved downside risk control and a stronger reward-to-drawdown profile, both of which are key considerations for risk-averse investors. \par 
It is important to note that, in this idealized evaluation setting where transaction costs are not considered, all methods, including lexicon-based approaches, exhibit relatively strong performance. This highlights a key limitation of much of the existing literature, where sentiment-driven portfolio strategies are evaluated under frictionless assumptions. Such settings can lead to overly optimistic conclusions that may not hold in real-world trading environments, where transaction costs can significantly affect performance.

\begin{table}[h!]
\centering
\caption{Performance metrics of the sentiment-based portfolios at no transaction cost. For each metric, the best performance is shown in bold, and the second-best is underlined.}
\label{tab:performance_0bps}
\resizebox{\linewidth}{!}{%
\begin{tabular}{lccccc}
\specialrule{.1em}{0.2em}{0.3em}
\textbf{Method} & \makecell{\textbf{Cumulative}\\\textbf{Return (\%)}} & \makecell{\textbf{Annualized}\\\textbf{Return (\%)}} & \textbf{Sharpe} & \textbf{Sortino} & \textbf{Calmar} \\
\specialrule{.1em}{0.2em}{0.3em}
\textbf{S\&P 500}  & 83.12 & 11.34 & 0.62 & 0.81 & 0.41 \\
\specialrule{.1em}{0.2em}{0.3em}
HIV-4        & 90.07  & 12.88  & 0.81 & 1.25 & 0.67 \\
VADER        & 82.81  & 11.76  & 0.75 & 1.21 & 0.34 \\
LMD          & 139.88 & 20.62  & 1.26 & 1.96 & 1.17 \\
FinBERT      & 199.19 & 29.64  & 1.65 & 2.39 & 1.24 \\
FinLlama     & \underline{260.74} & \underline{39.47}  & \underline{2.33} & \underline{3.48} & \underline{3.30} \\
\textbf{FinDPO (Ours)} & \textbf{747.10} & \textbf{111.78} & \textbf{3.41} & \textbf{6.05} & \textbf{11.94} \\
\specialrule{.1em}{0.2em}{0.3em}
\end{tabular}
}
\end{table}

\subsection{Incorporating Transaction Costs into Portfolio Construction}
We further assessed the impact of transaction costs on portfolio performance to better reflect practical trading conditions. In real-world settings, these costs can significantly reduce returns, especially for high-turnover strategies such as the daily rebalancing strategy we followed. Given that our investable universe consisted of relatively liquid instruments, specifically the S\&P 500 constituents, we varied the transaction cost parameter $k$ from 1 to 5 basis points (bps), which aligns with the typical 2–5 bps range observed in U.S. equities. To incorporate transaction costs, we adjusted daily portfolio returns using the standard linear model as
\begin{equation}
    r_t = R_t - k\cdot \text{Turnover}_t,
\end{equation}
where $R_t$ is the unadjusted portfolio return on day $t$, and 
\begin{equation}
    \text{Turnover}_t = \sum_i \left| w_{t}^{(i)} - w_{t-1} ^{(i)}\right|
\end{equation}
represents the total absolute change in portfolio weights across all assets, accounting for both long and short positions. \par 
\begin{table}[ht]
\centering
\caption{Profitability and risk-adjusted performance of the portfolios constructed by the six sentiment-based methods across different transaction cost levels. For all metrics, higher values indicate better performance. Positive (desired) values are highlighted in green and negative (undesired) values are highlighted in red. For each metric and cost level, the best performance is shown in bold, and the second-best is underlined.}
\label{tab:performance_by_cost}
\resizebox{\linewidth}{!}{%
\begin{tabular}{llccccc}
\specialrule{.1em}{0.2em}{0.3em}
\textbf{Cost (bps)} & \textbf{Method} & \makecell{\textbf{Cumulative}\\\textbf{Return (\%)}} & \makecell{\textbf{Annualized}\\\textbf{Return (\%)}} & \textbf{Sharpe} & \textbf{Sortino} & \textbf{Calmar} \\
\specialrule{.1em}{0.2em}{0.3em}
& \textbf{S\&P 500}  & 83.12 & 11.34 & 0.62 & 0.81 & 0.41 \\
\specialrule{.1em}{0.2em}{0.3em}
\multirow{6}{*}{\textbf{1.0}}
 & HIV-4        & \textcolor{ForestGreen}{36.17}  & \textcolor{ForestGreen}{4.41}   & \textcolor{ForestGreen}{0.28} & \textcolor{ForestGreen}{0.50} & \textcolor{ForestGreen}{0.21} \\
 & VADER       & \textcolor{ForestGreen}{29.32}  & \textcolor{ForestGreen}{3.37}   & \textcolor{ForestGreen}{0.21} & \textcolor{ForestGreen}{0.43} & \textcolor{ForestGreen}{0.09} \\
  & LMD         & \textcolor{ForestGreen}{85.54}  & \textcolor{ForestGreen}{12.09}  & \textcolor{ForestGreen}{0.74} & \textcolor{ForestGreen}{1.20} & \textcolor{ForestGreen}{0.52} \\
 & FinBERT     & \textcolor{ForestGreen}{144.46} & \textcolor{ForestGreen}{21.06}  & \textcolor{ForestGreen}{1.17} & \textcolor{ForestGreen}{1.74} & \textcolor{ForestGreen}{0.71} \\
 & FinLlama    & \textcolor{ForestGreen}{\underline{205.16}} & \textcolor{ForestGreen}{\underline{30.75}}  & \textcolor{ForestGreen}{\underline{1.81}} & \textcolor{ForestGreen}{\underline{2.75}} & \textcolor{ForestGreen}{\underline{2.05}} \\
 & \textbf{FinDPO (Ours)}       & \textbf{\textcolor{ForestGreen}{689.48}} & \textbf{\textcolor{ForestGreen}{102.76}} & \textbf{\textcolor{ForestGreen}{3.14}} & \textbf{\textcolor{ForestGreen}{5.61}} & \textbf{\textcolor{ForestGreen}{8.04}} \\
\specialrule{.1em}{0.2em}{0.3em}
\multirow{6}{*}{\textbf{2.0}}
 & HIV-4        & \textcolor{red}{-17.74} & \textcolor{red}{-4.05}  & \textcolor{red}{-0.26} & \textcolor{red}{-0.25} & \textcolor{red}{-0.07} \\
 & VADER       & \textcolor{red}{-24.16} & \textcolor{red}{-5.03}  & \textcolor{red}{-0.32} & \textcolor{red}{-0.35} & \textcolor{red}{-0.06} \\
  & LMD         & \textcolor{ForestGreen}{31.20}  & \textcolor{ForestGreen}{3.56}   & \textcolor{ForestGreen}{0.22} & \textcolor{ForestGreen}{0.44} & \textcolor{ForestGreen}{0.15} \\
 & FinBERT     & \textcolor{ForestGreen}{89.72}  & \textcolor{ForestGreen}{12.47}  & \textcolor{ForestGreen}{0.69} & \textcolor{ForestGreen}{1.08} & \textcolor{ForestGreen}{0.37} \\
 & FinLlama    & \textcolor{ForestGreen}{\underline{149.59}} & \textcolor{ForestGreen}{\underline{22.04}}  & \textcolor{ForestGreen}{\underline{1.30}} & \textcolor{ForestGreen}{\underline{2.00}} & \textcolor{ForestGreen}{\underline{1.03}} \\
 & \textbf{FinDPO (Ours)}       & \textbf{\textcolor{ForestGreen}{631.85}} & \textbf{\textcolor{ForestGreen}{93.73}}  & \textbf{\textcolor{ForestGreen}{2.86}} & \textbf{\textcolor{ForestGreen}{5.15}} & \textbf{\textcolor{ForestGreen}{5.47}} \\
\specialrule{.1em}{0.2em}{0.3em}
\multirow{6}{*}{\textbf{3.0}}
 & HIV-4        & \textcolor{red}{-71.64} & \textcolor{red}{-12.52} & \textcolor{red}{-0.79} & \textcolor{red}{-0.99} & \textcolor{red}{-0.17} \\
 & VADER       & \textcolor{red}{-77.64} & \textcolor{red}{-13.43} & \textcolor{red}{-0.85} & \textcolor{red}{-1.13} & \textcolor{red}{-0.16} \\
  & LMD         & \textcolor{red}{-23.14} & \textcolor{red}{-4.97}  & \textcolor{red}{-0.30} & \textcolor{red}{-0.32} & \textcolor{red}{-0.09} \\
 & FinBERT     & \textcolor{ForestGreen}{34.99}  & \textcolor{ForestGreen}{3.88}   & \textcolor{ForestGreen}{0.22} & \textcolor{ForestGreen}{0.42} & \textcolor{ForestGreen}{0.12} \\
 & FinLlama    & \textcolor{ForestGreen}{\underline{94.02}}  & \textcolor{ForestGreen}{\underline{13.32}}  & \textcolor{ForestGreen}{\underline{0.79}} & \textcolor{ForestGreen}{\underline{1.26}} & \textcolor{ForestGreen}{\underline{0.49}} \\
 & \textbf{FinDPO (Ours)}       & \textbf{\textcolor{ForestGreen}{574.22}} & \textbf{\textcolor{ForestGreen}{84.71}}  & \textbf{\textcolor{ForestGreen}{2.59}} & \textbf{\textcolor{ForestGreen}{4.69}} & \textbf{\textcolor{ForestGreen}{3.94}} \\
\specialrule{.1em}{0.2em}{0.3em}
\multirow{6}{*}{\textbf{4.0}}
 & HIV-4        & \textcolor{red}{-125.55}& \textcolor{red}{-20.99} & \textcolor{red}{-1.32} & \textcolor{red}{-1.73} & \textcolor{red}{-0.23} \\
 & VADER       & \textcolor{red}{-131.13}& \textcolor{red}{-21.84} & \textcolor{red}{-1.39} & \textcolor{red}{-1.90} & \textcolor{red}{-0.23} \\
  & LMD         & \textcolor{red}{-77.49} & \textcolor{red}{-13.51} & \textcolor{red}{-0.83} & \textcolor{red}{-1.08} & \textcolor{red}{-0.19} \\
 & FinBERT     & \textcolor{red}{-19.74} & \textcolor{red}{-4.72}  & \textcolor{red}{-0.26} & \textcolor{red}{-0.24} & \textcolor{red}{-0.06} \\
 & FinLlama    & \textcolor{ForestGreen}{\underline{38.44}}  & \textcolor{ForestGreen}{\underline{4.60}}   & \textcolor{ForestGreen}{\underline{0.27}} & \textcolor{ForestGreen}{\underline{0.52}} & \textcolor{ForestGreen}{\underline{0.16}} \\
 & FinDPO      & \textbf{\textcolor{ForestGreen}{516.59}} & \textbf{\textcolor{ForestGreen}{75.68}}  & \textbf{\textcolor{ForestGreen}{2.31}} & \textbf{\textcolor{ForestGreen}{4.22}} & \textbf{\textcolor{ForestGreen}{2.92}} \\
\specialrule{.1em}{0.2em}{0.3em}
\multirow{6}{*}{\textbf{5.0}}
 & HIV-4        & \textcolor{red}{-179.46}& \textcolor{red}{-29.46} & \textcolor{red}{-1.85} & \textcolor{red}{-2.46} & \textcolor{red}{-0.28} \\
 & VADER       & \textcolor{red}{-184.61}& \textcolor{red}{-30.24} & \textcolor{red}{-1.92} & \textcolor{red}{-2.66} & \textcolor{red}{-0.29} \\
  & LMD         & \textcolor{red}{-131.83}& \textcolor{red}{-22.05} & \textcolor{red}{-1.35} & \textcolor{red}{-1.84} & \textcolor{red}{-0.25} \\
 & FinBERT     & \textcolor{red}{-74.48} & \textcolor{red}{-13.31} & \textcolor{red}{-0.74} & \textcolor{red}{-0.90} & \textcolor{red}{-0.18} \\
 & FinLlama    & \textcolor{red}{\underline{-17.13}} & \textcolor{red}{\underline{-4.13}}  & \textcolor{red}{\underline{-0.24}} & \textcolor{red}{\underline{-0.23}} & \textcolor{red}{\underline{-0.06}} \\
 & \textbf{FinDPO (Ours)}       & \textbf{\textcolor{ForestGreen}{458.97}} & \textbf{\textcolor{ForestGreen}{66.64}}  & \textbf{\textcolor{ForestGreen}{2.03}} & \textbf{\textcolor{ForestGreen}{3.75}} & \textbf{\textcolor{ForestGreen}{2.21}} \\
\specialrule{.1em}{0.2em}{0.3em}
\end{tabular}
}
\end{table}
When transaction costs are considered, a clear performance gap emerges between FinDPO and all other sentiment-based portfolio construction methods. Even at a high transaction cost of 5 bps, FinDPO remained the only method that consistently delivered significant positive returns and robust risk-adjusted performance, achieving an annualized return of 67\% with a Sharpe ratio of 2.0. In contrast, all other methods, including FinLlama, exhibited substantial degradation in both absolute and risk-adjusted performance, with all yielding very low to highly negative cumulative returns at realistic transaction cost levels of 4–5 bps. The inability of these methods to maintain profitability under transaction costs highlights their sensitivity to turnover and the lack of robustness in real-world trading environments. \par 
In contrast, FinDPO demonstrates strong and stable performance across all transaction cost levels, underscoring its effectiveness at extracting robust trading signals and its resilience to market frictions. These results highlight that FinDPO is not only highly profitable under idealized conditions but also realistically deployable, maintaining high annualized returns and strong risk-adjusted performance even under challenging market conditions.
\section{Conclusion}
We have introduced an innovative approach to financial sentiment analysis which rests upon Direct Preference Optimization (DPO), a post-training alignment technique based on human preference optimization.  By leveraging the expressive power and contextual understanding of a pre-trained LLM and adapting it to financial sentiment analysis, this approach has enabled the model to become more attuned to the nuanced language of the financial domain, while also minimizing resource utilization and computational overhead.
Owing to the strong generalization capability of preference-based alignment techniques compared to SFT approaches, including instruction tuning, our proposed FinDPO model has achieved a weighted F1 score that outperforms the SOTA FinGPT v3.3 model by 11\% on the average. \par 
Uniquely, our framework has enabled fine-tuned causal LLMs, which have shown superior performance on machine learning benchmarks, to be integrated into portfolio optimization strategies. This has allowed for the evaluation of our FinDPO model using real-world financial metrics, which extend beyond the traditional classification-focused evaluations currently found in the literature. Overall, these advantages have established FinDPO, to the best of our knowledge, as the first sentiment-based method to maintain substantial positive returns and strong risk-adjusted performance even under high transaction costs, thus making it robust and practically deployable in real-world trading environments. \par 
\textbf{Disclaimer: Nothing herein is financial advice, and NOT a recommendation to trade real money. Please use common sense and always first consult a professional before trading or investing.}

  

\begin{acks}
The authors would like to thank Thanos Konstantinidis, Portfolio Manager at Engineers Gate, for their valuable feedback and suggestions regarding portfolio construction and the calculation of transaction costs.
\end{acks}

\bibliographystyle{ACM-Reference-Format}
\bibliography{references}

\appendix

\end{document}